# An HMM Based Named Entity Recognition System for Indian Languages: The JU System at ICON 2013

**Vivekananda Gayen**
Computer Science & Technology Dept.
Central Calcutta Polytechnic
Kolkata – 700014, India
`vivek3gayen@gmail.com`

**Kamal Sarkar**
Computer Science & Engineering Dept.
Jadavpur University
Kolkata-700032, India
`jukamal2001@yahoo.com`

## Abstract

This paper reports about our work in the ICON 2013 NLP TOOLS CONTEST on Named Entity Recognition. We submitted runs for Bengali, English, Hindi, Marathi, Punjabi, Tamil and Telugu. A statistical HMM (Hidden Markov Models) based model has been used to implement our system. The system has been trained and tested on the NLP TOOLS CONTEST: ICON 2013 datasets. Our system obtains F-measures of 0.8599, 0.7704, 0.7520, 0.4289, 0.5455, 0.4466, and 0.4003 for Bengali, English, Hindi, Marathi, Punjabi, Tamil and Telugu respectively.

## 1 Introduction

Named entity recognition involves locating and classifying the names in text. The objective of named entity recognition is to identify and classify every word/term in a document into some predefined categories like person name, location name, organization name, miscellaneous name (date, time, percentage and monetary expressions etc.) and "none-of-the-above".

NER is an important task, having applications in Information Extraction, Question Answering, Machine Translation, Summarization and other NLP applications. This paper presents a HMM (Hidden Markov Model) based NER system for Indian languages which is designed for the ICON 2013 NLP tool contest, the goal of which is to perform NE recognition on a variety of types: artifact, entertainment, facilities, location, locomotive, materials, organisms, organization, person, plants, count, distance, money, quantity, date, day, period, time and year.

The ICON 2013 NLP tool contest was defined to build the NER systems for seven Indian languages - Bengali, English, Hindi, Marathi, Punjabi, Tamil and Telugu for which training data, development data and test data were provided. Since our HMM based system is of language independent nature. We have participated for all seven languages. We have not used gazetteers for these tasks because gazetteer lists for all seven languages were not available with us.

The earliest works on named entity recognition used a variety of techniques for named entity recognition (NER). The two major approaches to NER are: Rule based (Linguistic) approaches and Machine Learning (ML) based approaches.

The rule based approaches typically use rules manually written by human (Grishman, 1995; McDonald, 1996; Wakao et al., 1996).

Machine learning (ML) based techniques for NER make use of a large amount of NE annotated training data to acquire high level language knowledge. Several ML techniques have been successfully used for the NER tasks such as Markov Model (HMM) (Bikel et al., 1997), Maximum Entropy (MaxEnt) (Borthwick, 1999; Kumar and Bhattacharyya, 2006), Conditional Random Field (CRF) (Li and Mccallum, 2004) etc.

Combinations of different ML approaches are also used. Srihari et al.(2000) combines MaxEnt,

Hidden Markov Model (HMM) and handcrafted rules to build an NER system.

NER systems also use gazetteer lists for identifying names. Both the linguistic approach (Grishman, 1995; Wakao et al., 1996) and the ML based approach (Borthwick, 1999; Srihari et al., 2000) may use gazetteer lists.

The NER tasks for Hindi have been presented in (Cucerzan and Yarowsky, 1999; Li and Mccallum, 2004; Ekbal et.al., 2009).

A discussion on the training data is given in Section 2. The HMM based NER system is described in Section 3. Various features used in NER are then discussed. Next we present the experimental results and related discussions in Section 4. Finally Section 5 concludes the paper.

## 2 Training Data

The data used for the training of our systems was provided. The annotated data uses Shakti Standard Format (SSF). Our system converts the SSF format data into the IOB format before training and the data converted in IOB format is used for training. IOB format uses a B−XXX tag that indicates the first word of an entity type XXX and I−XXX that is used for subsequent words of an entity. The tag "O" indicates the word is outside of an NE (i.e., not a part of a named entity). An additional tag E-XXX (this is not the part of IOB format) is also added to the tagset to tag the last word of an entity type XXX.

At the time of system development we have observed that the data (training, development and test), provided by the organizers of ICON 2013 NLP tool contest, contains several types of errors such as ambiguous POS tags (XC:?, QO:? Etc.), errors in NE tagging. These errors in the training corpora affects badly to the machine learning (ML) based models. But we have not made corrections of the errors in the training corpora at the time of our system development. All the results shown in the paper are obtained using the provided corpora without any modification in NE annotation.

## 3 HMM based Named Entity Tagging

A named entity recognizer based on Hidden Markov Model (HMM) assigns the best sequence of NE tags $t_1^n$ that is optimal for a given observation sequence $o_1^n$. The tagging problem becomes equivalent to searching for $\arg\max_{t_1^n} P(o_1^n | t_1^n) P(t_1^n)$ (by the application of Bayes' law), that is, we need to compute:

$$\hat{t}_1^n = \arg\max_{t_1^n} P(o_1^n | t_1^n) P(t_1^n) \qquad (1).$$

Where $t_1^n$ is a tag sequence and $o_1^n$ is an observation sequence, $P(t_1^n)$ is the prior probability of the tag sequence and $P(o_1^n | t_1^n)$ is the likelihood of the word sequence.

In general, HMM based sequence labeling tasks such as POS tagging use words in a sentence as an observation sequence (Sarkar and Gayen, 2012, 2013). But, the data released for ICON 2013 NLP tool contest is POS tagged and chunked, that is, some additional information such as POS tag and chunk tag for each word in a sentence is also available. To use this additional information for named entity recognition task, we consider a triplet:

<word, POS-tag, chunk-tag > as an observed symbol, that is, for a sentence of *n* words, the corresponding observation sequence will be as follows:

(<word$_1$, POS-tag$_1$, chunk-tag$_1$>, <word$_2$, POS-tag$_2$, chunk-tag$_2$>, <word$_3$, POS-tag$_3$, chunk-tag$_3$>, .........., <word$_n$, POS-tag$_n$, chunk-tag$_n$>) . Here an observation symbol o$_i$ corresponds to <word$_i$, POS-tag$_i$, chunk-tag$_i$>.

Equation (1) is too hard to compute directly. HMM taggers make Markov assumption which states that the probability of a tag is dependent only on a small, fixed number of previous tags. A bigram tagger considers that the probability of a tag depends only on the previous tag. For our proposed trigram model, the probability of a tag depends on two previous tags and $P(t_1^n)$ is computed as:

$$P(t_1^n) \approx \prod_{i=1}^{n} P(t_i | t_{i-1}, t_{i-2}) \qquad (2)$$

Depending on the assumption that the probability of a word appearing is dependent only on its own tag, $P(o_1^n | t_1^n)$ can be simplified to:

$$P(o_1^n | t_1^n) \approx \prod_{i=1}^{n} P(o_i | t_i) \qquad (3)$$

Plugging the above mentioned two equations (2) and (3) into (1) results in the following equation by which a bigram tagger estimates the most probable tag sequence:

$$\hat{t}_1^n = \arg\max_{t_1^n} P(t_1^n | o_1^n) P(t_1^n) \approx \arg\max_{t_1^n} \prod_{i=1}^{n} P(o_i | t_i) P(t_i | t_{i-1}) \quad (4)$$

Where: the tag transition probabilities, $P(t_i | t_{i-1})$, represent the probability of a tag given the previous tag. $P(o_i | t_i)$ represents the probability of an observed symbol given a tag.

Considering a special tag $t_{n+1}$ to indicate the end sentence boundary and two special tags $t_{-1}$ and $t_0$ at the starting boundary of the sentence and adding these three special tags to the tag set (Brants, 2000), gives the following equation for NE tagging:

$$\hat{t}_1^n = \arg\max_{t_1^n} P(t_1^n | o_1^n) P(t_1^n) \approx$$
$$\arg\max_{t_1^n} [\prod_{i=1}^{n} P(o_i | t_i) P(t_i | t_{i-1}, t_{i-2})] P(t_{n+1} | t_n) \quad (5)$$

The equation (5) is still computationally expensive because we need to consider all possible tag sequence of length *n*. So, dynamic programming approach is used to compute the equation (5).

At the training phase of HMM based NE tagging, observation probability matrix and tag transition probability matrix are created.

### 3.1 Computing Tag Transition Probabilities

As we can see from the equation (4) to find the most likely tag sequence for an observation sequence, we need to compute two kinds of probabilities: tag transition probabilities and word likelihoods or observation probabilities.

Our developed trigram HMM tagger requires to compute tag trigram probability, $P(t_i | t_{i-1}, t_{i-2})$, which is computed by the maximum likelihood estimate from tag trigram counts. To overcome the data sparseness problem, tag trigram probability is smoothed based on the bigram and unigram probabilities using the following equation:

$$P(t_i | t_{i-1}, t_{i-2}) = \lambda_1 \hat{P}(t_i | t_{i-1}, t_{i-2}) + \lambda_2 \hat{P}(t_i | t_{i-1}) + \lambda_3 \hat{P}(t_i) \quad (6)$$

$\hat{P}(t_i | t_{i-1}, t_{i-2})$, $\hat{P}(t_i | t_{i-1})$ and $\hat{P}(t_i)$ are the maximum likelihood estimates from counts for tag trigram, tag bigram and tag unigram respectively:

$$\hat{P}(t_i | t_{i-1}, t_{i-2}) = \frac{C(t_{i-2}, t_{i-1}, t_i)}{C(t_{i-2}, t_{i-1})}, \quad \hat{P}(t_i | t_{i-1}) = \frac{C(t_{i-1}, t_i)}{C(t_{i-1})},$$

$$\hat{P}(t_i) = \frac{C(t_i)}{N}$$

Where: $C(t_{i-2}, t_{i-1}, t_i)$ indicates the count of the tag sequence $<t_{i-2}, t_{i-1}, t_i>$ and $\lambda_1, \lambda_2, \lambda_3$ ($\lambda_1 + \lambda_2 + \lambda_3 = 1$) are the weights for the maximum likelihood estimates of trigram, bigram and unigram tag probabilities respectively computed based on corpus statistics. The values of the parameters: $\lambda_1, \lambda_2, \lambda_3$ are estimated using a smoothing technique called the deleted interpolation proposed in (Brants, 2000).

### 3.2 Computing Observation Probabilities

The observation probability of a observed triplet <word, POS-tag, chunk-tag >, which is the observed symbol in our case, is computed using the following equation (Sarkar and Gayen, 2012, 2013):

$$P(o | t) = \frac{C(o,t)}{C(o)} \quad (7)$$

### 3.3 Viterbi Decoding

The task of a decoder is to find the best hidden state sequence given an input HMM and a sequence of observations.

The Viterbi algorithm is the most common decoding algorithm used for HMM based tagging task. This is a standard application of the classic dynamic programming algorithm (Jurafsky and Martin, 2002). The Viterbi algorithm that we use, takes as input a single HMM and a set of observed sequence $O = (o_1 o_2 o_3 ... o_t)$ and returns the most probable state sequence, $Q = (q_1 q_2 q_3 ... q_t)$, together with its probability.

Given a tag transition probability matrix and the observation probability matrix, Viterbi decoding (used at the testing phase) accepts a text document in Indian language and finds the most likely tag sequence for each POS-tagged chunked sentence in the input document. Here a sentence is also submitted to the viterbi as the observation sequence of triplets:

(<word$_1$, POS-tag$_1$, chunk-tag$_1$>, <word$_2$, POS-tag$_2$, chunk-tag$_2$>, <word$_3$, POS-tag$_3$, chunk-tag$_3$>, ..., <word$_n$, POS-tag$_n$, chunk-tag$_n$>). After assigning the tag sequence to the observation sequence as mentioned above, POS-tag and chunk-tag information are removed from the output and thus the out-put for an input sentence is converted to a NE-tagged sentence.

We have used the Viterbi algorithm presented in (Jurafsky and Martin, 2002) for finding the most likely tag sequence for a given observation sequence.

One of the important problems to apply Viterbi decoding algorithm is how to handle unknown triplets in the input. The unknown triplets are triplets which are not present in the training set and hence their observation probabilities are not known. To handle this problem, we estimate the observation probability of an unknown one by analyzing POS-tag, chunk-tag and the suffix of the word associated with the corresponding the triplet. We estimate the observation probability of an unknown observed triplet in the following ways:

The observation probabilities of unknown triplet <word, POS-tag, chunk-tag > corresponding to a word in the input sentence are decided according to the suffix of a pseudo word formed by adding POS-tag and chunk-tag to the end of the word. We find the observation probabilities of such unknown pseudo words using suffix analysis of all rare pseudo words (frequency <=2) in the training corpus for the concerned language. Since unknown pseudo words are infrequent and using suffixes of infrequent pseudo words in the lexicon is a better approximation for unknown pseudo words (Brants, 2000). The term suffix as used in this context means "a sequence of characters occurring at the end of a pseudo word" which is not necessarily a linguistically meaningful suffix. The maximum length of suffix has been tuned on the development set for the corresponding languages: suffix length is set to 8 for Bengali, 9 for English, 9 for Hindi, 9 for Marathi, 9 for Punjabi, 16 for Tamil and 13 for Telugu respectively. The probability of a tag given a suffix of length i is computed as: $P(t \mid suffix\text{-}of\text{-}len(i))$. These probabilities are smoothed using successively shorter and shorter suffixes (Brants, 2000). This can be formulated in recursive way as:

$$P(t \mid suffix-of-len(i)) = \frac{\hat{p}(t \mid suffix-of-len(i)) + \theta_i p(t \mid suffix-of-len(i))}{1+\theta_i} \quad (8)$$

Where: $\hat{p}$ is the maximum likelihood probability based on the count of <tag, suffix> pair in all rare pseudo words (frequency <=2) in the corpus.

All $\theta_i$ are set to the standard deviation of the unconditioned maximum likelihood probabilities ($\hat{p}(t_i)$) of the tags in the training corpus (Brants, 2000). $P(t \mid suffix\text{-}of\text{-}len(i))$ gives an estimate of $P(t_i \mid w_i)$. But for HMM based tagging we need to compute the likelihood $P(w_i \mid t_i)$ which is computed from $P(t_i \mid w_i)$ using Bayesian inversion that uses Bayes rule and prior $P(t_i)$.

## 4 Evaluation and Results

After getting the NE-tagged output in IOB format from the HMM model, we observed that the tagged output contains some occurrences of a sequence of I-XXXs where the left boundary of each such sequence is a transition from the tag "O" to I-XXXs (according to the IOB format, the left boundary of a named entity is a transition from any tag to B-XXX). We have also observed that the word sequence to which this type of tag sequence is assigned is not really a named entity. So, considering this as the errors of the model, we replace such a sequence of I-XXXs in the output by a sequence of "o". After doing this post-processing on the output produced by the HMM model, the final output file is produced.

Our developed NER system has been evaluated using the traditional precision, recall and F-measure. For training, tuning and testing our system, we have used the datasets for 7 Indian languages, released by the organizers of the ICON 2013 NLP tool contest.

We train separately our developed named entity recognizer on the training data for each of the languages and tune the parameters of our system on the development data for the language under consideration. After learning the tuning parameters, we test our system on the test data for the concerned language.

The description of the data for each of 7 Indian languages is shown in the Table1

| Lan- | Total tokens | | | NE |
| guage | Train-ing | Devel-opment | Test | Types |
| --- | --- | --- | --- | --- |
| Bengali | 43732 | 6116 | 5938 | 24 |
| English | 91869 | 15839 | 14438 | 21 |
| Hindi | 68608 | 10678 | 8931 | 22 |
| Marathi | 72628 | 8975 | 7871 | 21 |
| Punjabi | 63253 | 8381 | 8008 | 21 |
| Tamil | 74077 | 7160 | 6608 | 25 |
| Telugu | 34910 | 6014 | 4288 | 22 |

Table1. The description of the data for each of 7 Indian languages

| NER Systems | Precision | Recall | F-measure |
|---|---|---|---|
| Bengali | 0.8447 | 0.8756 | 0.8599 |
| English | 0.7683 | 0.7725 | 0.7704 |
| Hindi | 0.7540 | 0.7500 | 0.7520 |
| Marathi | 0.5305 | 0.3600 | 0.4289 |
| Punjabi | 0.5497 | 0.5413 | 0.5455 |
| Tamil | 0.3221 | 0.7279 | 0.4466 |
| Telugu | 0.3773 | 0.4263 | 0.4003 |

Table2. Performance of our named entity recognition system on the test data for Bengali, English, Hindi, Marathi, Punjabi, Tamil and Telugu.

Table 2 shows the performances of our developed named entity recognition system on the test data for 7 Indian Languages namely Bengali, English, Hindi, Marathi, Punjabi, Tamil and Telugu.

## 5 Conclusion

This paper describes a named entity recognition system for Indian Languages namely Bengali, English, Hindi, Marathi, Punjabi, Tamil and Telugu. The named entity recognition system has been developed using Visual Basic platform so that a suitable user interface can be designed for the novice users. The system has been designed in such a way that only changing the training corpus in a file can make the system portable to a new Indian language.

## References

Andrew Borthwick. 1999. *A Maximum Entropy Approach to Named Entity Recognition*. Ph.D. thesis, Computer Science Department, New York University.

Asif Ekbal, Rejwanul Haque, Amitava Das, Venkateswarlu Poka and Sivaji Bandyopadhyay. 2008. *Language Independent Named Entity Recognition in Indian Languages.* In Proceedings of IJCNLP workshop on NERSSEAL.

Asif Ekbal and Sivaji Bandyopadhyay. 2009. *A conditional random field approach for named entity recognition in Bengali and Hindi*. Linguistic Issues in Language Technology, 2(1).

Daniel Jurafsky and James H. Martin. 2002. *"Speech and Language Processing An Intoduction to Natural Language Processing, Computational Linguistics and Speech Recognition"*, Preason Education Series.

Daniel M. Bikel, Scott Miller, Richard Schwartz and Ralph Weischedel. 1997. Nymble: *A High Performance Learning Name-finder*. In Proceedings of the Fifth Conference on Applied Natural Language Processing, 194–201.

David D. McDonald. 1996. *Internal and external evidence in the identification and semantic categorization of proper names.* In B. Boguraev and J. Pustejovsky, editors, Corpus Processing for Lexical Acquisition, 21–39.

Kamal Sarkar and Vivekananda Gayen. 2012. *A practical part-of-speech tagger for Bengali*. In Proceedings of the third International conference on Emerging Applications of Information Technology (EAIT), Kolkata. pp. 36-40.

Kamal Sarkar and Vivekananda Gayen. 2013 . *A Trigram HMM-Based POS Tagger for Indian Languages*. In proceedings of International Conference on Frontiers of Intelligent Computing: Theory and Applications (FICTA). pp. 205-212.

Karthik Gali, Harshit Surana, Ashwini Vaidya, Praneeth Shishtla and Dipti M. Sharma. 2008. *Aggregating Machine Learning and Rule Based Heuristics for Named Entity Recgnition*. In Proceedings of IJCNLP workshop on NERSSEAL.

N. Kumar and Pushpak Bhattacharyya. 2006. *Named Entity Recognition in Hindi using MEMM*. In Technical Report, IIT Bombay, India.

Praveen Kumar P. and Ravi Kiran V. 2008. *Hybrid Named Entity Recognition System for South-South East Indian Languages.* InProceedings of IJCNLP workshop on NERSSEAL.

Ralph Grishman. 1995. *The New York University System MUC-6 or Where's the syntax?* In Proceedings of the Sixth Message Understanding Conference.

Rohini Srihari, Cheng Niu and Wei Li. 2000. *A Hybrid Approach for Named Entity and Sub-Type Tagging*. In Proceedings of the sixth conference on Applied natural language processing.

Silviu Cucerzan and David Yarowsky. 1999. *Language Independent Named Entity Recognition Combining Morphological and Contextual Evidence.* In Proceedings of the Joint SIGDAT Conference on EMNLP and VLC 1999, 90–99.

Takahiro Wakao, Robert Gaizauskas and Yorick Wilks. 1996. *Evaluationof an algorithm for the recognition and classification of proper names.* In Proceedings of COLING-96.

Thorsten Brants. 2000. *TnT – A statistical part-of-speech tagger*. In proceedings of the 6[th] Applied NLP Conference, pp. 224-231.

Wei Li and Andrew McCallum. 2004. *Rapid Development of Hindi Named Entity Recognition using Conditional Random Fields and Feature Induction (Short Paper)*. In ACM Transactions on Computational Logic.